\documentclass[conference]{IEEEtran}

\usepackage{times}
\usepackage[numbers]{natbib}
\usepackage{multirow}
\usepackage[hyperref]{utils}

\graphicspath{{figures/}}
\setenumerate{leftmargin=*}
\setitemize{leftmargin=*}

\title{Adaptive-Control-Oriented Meta-Learning\\for Nonlinear Systems\vspace{-0.3em}}

\author{%
    \IEEEauthorblockN{%
        Spencer M.~Richards\IEEEauthorrefmark{1},
        Navid Azizan\IEEEauthorrefmark{1}\IEEEauthorrefmark{2},
        Jean-Jacques Slotine\IEEEauthorrefmark{2},
        and Marco Pavone\IEEEauthorrefmark{1}
    }\vspace{0.3em}
    \IEEEauthorblockA{%
        \IEEEauthorrefmark{1}
        Department of Aeronautics \& Astronautics,
        Stanford University, California, U.S.A.
        \\
        \IEEEauthorrefmark{2}
        Department of Mechanical Engineering,
        Massachusetts Institute of Technology, Massachusetts, U.S.A.
        \\[0.3em]
        Email:
        \IEEEauthorrefmark{1}\texttt{\{spenrich,pavone\}@stanford.edu},
        \IEEEauthorrefmark{2}\texttt{\{azizan,jjs\}@mit.edu}
    }\vspace{-2em}
}

\begin{document}

\maketitle

\begin{abstract}
    Real-time adaptation is imperative to the control of robots operating in complex, dynamic environments. Adaptive control laws can endow even nonlinear systems with good trajectory tracking performance, provided that any uncertain dynamics terms are linearly parameterizable with known nonlinear features. However, it is often difficult to specify such features a priori, such as for aerodynamic disturbances on rotorcraft or interaction forces between a manipulator arm and various objects. In this paper, we turn to data-driven modeling with neural networks to learn, offline from past data, an adaptive controller with an internal parametric model of these nonlinear features. Our key insight is that we can better prepare the controller for deployment with control-oriented meta-learning of features in closed-loop simulation, rather than regression-oriented meta-learning of features to fit input-output data. Specifically, we meta-learn the adaptive controller with closed-loop tracking simulation as the base-learner and the average tracking error as the meta-objective. With a nonlinear planar rotorcraft subject to wind, we demonstrate that our adaptive controller outperforms other controllers trained with regression-oriented meta-learning when deployed in closed-loop for trajectory tracking control.
\end{abstract}

\IEEEpeerreviewmaketitle

\begin{figure}[t]
    \centering
    \includegraphics[width=\columnwidth]{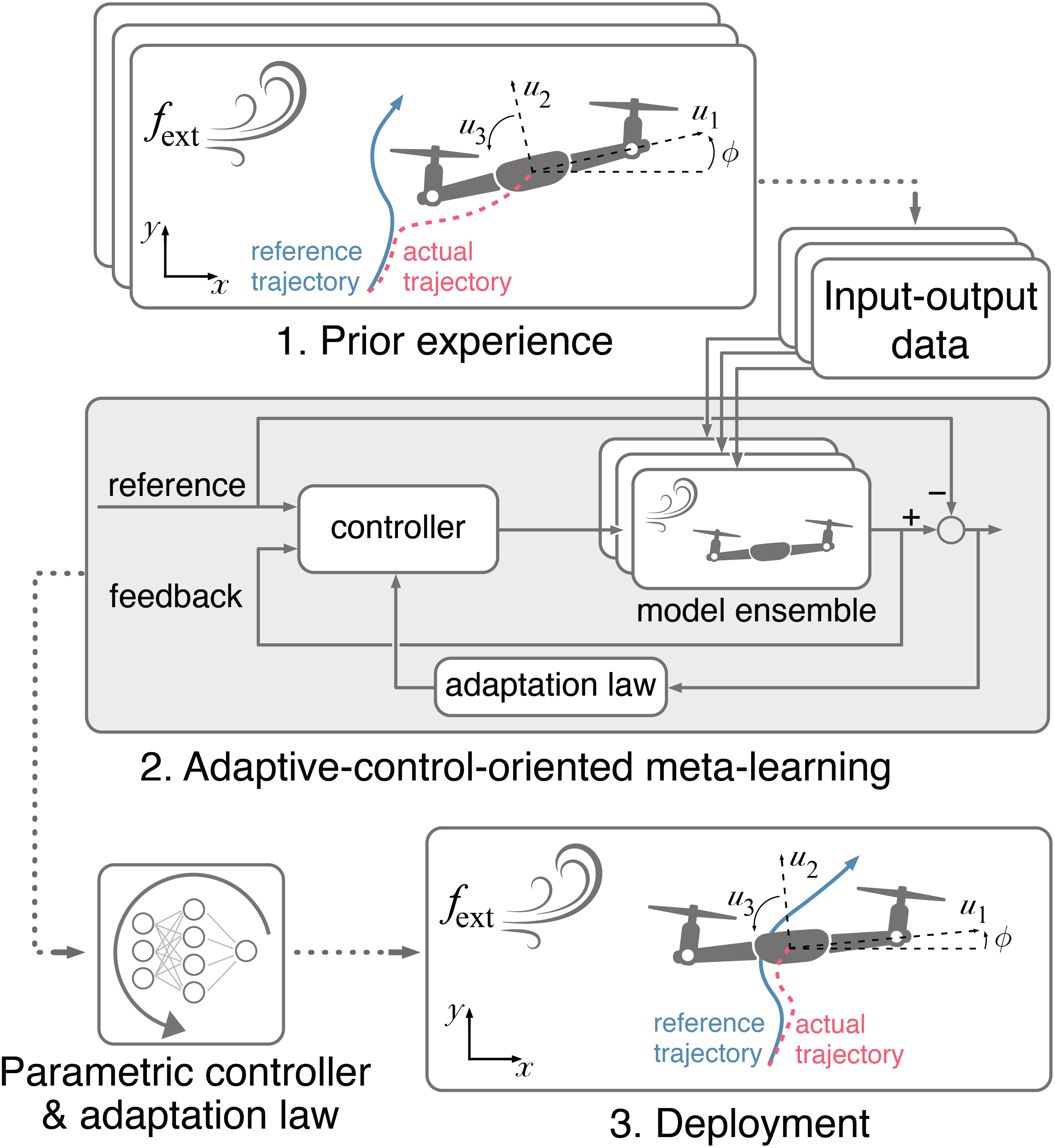}
    \vspace{-1.5em}
    \caption{%
        While roboticists can often derive a model for how control inputs affect the system state, it is much more difficult to model prevalent external forces (\eg from aerodynamics, contact, and friction) that adversely affect tracking performance. In this work, we present a method to meta-learn an adaptive controller offline from previously collected data. Our meta-learning is \emph{control-oriented} rather than regression-oriented; specifically, we: 1) collect input-output data on the true system, 2) train a parametric adaptive controller in closed-loop simulation to adapt well to each model of an ensemble constructed from past input-output data, and 3) test our adaptive controller on the real system. Our method contextualizes training within the downstream control objective, thereby engendering good tracking results at test time, which we demonstrate on a Planar Fully-Actuated Rotorcraft (PFAR) subject to wind.
    }\label{fig:pfar}
    \vspace{-1.5em}
\end{figure}

\section{Introduction}

Performant control in robotics is impeded by the complexity of the dynamical system consisting of the robot itself (\ie its nonlinear equations of motion) and the interactions with its environment. Roboticists can often derive a physics-based robot model, and then choose from a suite of nonlinear control laws that each offer desirable control-theoretic properties (\eg good tracking performance) in known, simple environments. Even in the face of model uncertainty, nonlinear control can still yield such properties with the help of \emph{real-time adaptation} to \emph{online} measurements, provided the uncertainty enters the system in a known, structured manner.

However, when a robot is deployed in complex scenarios, it is generally intractable to know even the structure of all possible configurations and interactions that the robot may experience. To address this, system identification and data-driven control seek to learn an accurate input-output model from past measurements. Recent years have also seen a dramatic proliferation of research in machine learning for control by leveraging powerful approximation architectures to predict and optimize the behaviour of dynamical systems. In general, such rich models require extensive data and computation to back-propagate gradients for many layers of parameters, and thus usually cannot be used in fast nonlinear control loops.

Moreover, machine learning of dynamical system models often prioritizes fitting input-output data, \ie it is \emph{regression-oriented}, with the rationale that designing a controller for a highly accurate model engenders better closed-loop performance on the real system. However, decades of work in system identification and adaptive control recognize that, since a model is often learned for the purpose of control, the learning process itself should be tailored to the \emph{downstream control objective}. This concept of \emph{control-oriented} learning is exemplified by fundamental results in adaptive control theory for linearly parameterizable systems; guarantees on tracking convergence can be attained \emph{without} convergence of the parameter estimates to those of the true system. 

\paragraph*{Contributions} In this work, we acknowledge this distinction between regression-oriented and control-oriented learning, and propose a control-oriented method to learn a parametric adaptive controller that performs well in closed-loop at test time. Critically, our method (outlined in \cref{fig:pfar}) focuses on \emph{offline} learning from past trajectory data. We formalize training the adaptive controller as a semi-supervised, bi-level meta-learning problem, with the average integrated tracking error across chosen reference trajectories as the meta-objective. We use a closed-loop simulation with our adaptive controller as a base-learner, which we then back-propagate gradients through. We discuss how our formulation can be applied to adaptive controllers for general dynamical systems, then specialize it to the case of nonlinear mechanical systems. Through our experiments, we show that by injecting the downstream control objective into offline meta-learning of an adaptive controller, we improve closed-loop trajectory tracking performance at test time in the presence of widely varying disturbances. We provide code to reproduce our results at \url{https://github.com/StanfordASL/Adaptive-Control-Oriented-Meta-Learning}.

\section{Related Work}\vspace{-0.1em}

In this section, we review three key areas of work related to this paper: control-oriented system identification, adaptive control, and meta-learning.\vspace{-0.1em}

\subsection{Control-Oriented System Identification}
Learning a system model for the express purpose of closed-loop control has been a hallmark of linear system identification since at least the early 1970s~\cite{AstromWittenmark1971}. Due to the sheer amount of literature in this area, we direct readers to \citet{Ljung1999} and \citet{Gevers2005}. Some salient works are the demonstrations by \citet{Skelton1989} on how large open-loop modelling errors do not necessarily cause poor closed-loop prediction, and the theory and practice from \citet{HjalmarssonGeversEtAl1996} and \citet{ForssellLjung2000} for iterative online closed-loop experiments that encourage convergence to a model with optimal closed-loop behaviour. In this paper, we focus on \emph{offline meta-learning} targeting a downstream closed-loop control objective, to train adaptive controllers for \emph{nonlinear} systems.

In nonlinear system identification, there is an emerging body of literature on data-driven, constrained learning for dynamical systems that encourages learned models and controllers to perform well in closed-loop. \citet{Khansari-ZadehBillard2011} and \citet{MedinaBillard2017} train controllers to imitate known invertible dynamical systems while constraining the closed-loop system to be stable. \citet{ChangRoohiEtAl2019} and \citet{SunJhaEtAl2020} jointly learn a controller and a stability certificate for known dynamics to encourage good performance in the resulting closed-loop system. \citet{SinghRichardsEtAl2020} jointly learn a dynamics model and a stabilizability certificate that regularizes the model to perform well in closed-loop, even with a controller designed a posteriori. Overall, these works concern learning a fixed model-controller pair. Instead, with offline meta-learning, we train an \emph{adaptive} controller that can update its internal representation of the dynamics online. We discuss future work explicitly incorporating stability constraints in~\cref{sec:conclusion}.

\subsection{Adaptive Control}

Broadly speaking, adaptive control concerns parametric controllers paired with an \emph{adaptation law} that dictates how the parameters are adjusted online in response to signals in a dynamical system \cite{SlotineLi1991,NarendraAnnaswamy2005,LandauLozanoEtAl2011,IoannouSun2012}. Since at least the 1950s, researchers in adaptive control have focused on parameter adaptation prioritizing control performance over parameter identification \cite{AseltineManciniEtAl1958}. Indeed, one of the oldest adaptation laws, the so-called MIT rule, is essentially gradient descent on the integrated squared \emph{tracking} error \cite{MareelsAndersonEtAl1987}. Tracking convergence to a reference signal is the primary result in Lyapunov stability analyses of adaptive control designs~\cite{NarendraValavani1978,NarendraValavani1980}, with parameter convergence as a secondary result if the reference is persistently exciting~\cite{AndersonJohnson1982,BoydSastry1986}. In the absence of persistent excitation, \citet{BoffiSlotine2021} show certain adaptive controllers also ``implicitly regularize'' \cite{AzizanHassibi2019,AzizanLaleEtAl2019} the learned parameters to have small Euclidean norm; moreover, different forms of implicit regularization (\eg sparsity-promoting) can be achieved by certain modifications of the adaptation law. Overall, adaptive control prioritizes control performance while learning parameters on a ``need-to-know'' basis, which is a principle that can be extended to many learning-based control contexts~\cite{WensingSlotine2020}.

Stable adaptive control of nonlinear systems often relies on linearly parameterizable dynamics with known nonlinear basis functions, \ie \emph{features}, and the ability to cancel these nonlinearities stably with the control input when the parameters are known exactly \cite{SlotineLi1987,SlotineLi1989,SlotineLi1991,LopezSlotine2021}. When such features cannot be derived a priori, function approximators such as neural networks~\cite{SannerSlotine1992,JoshiChowdhary2019,JoshiVirdiEtAl2020} and Gaussian processes~\cite{GrandeChowdharyEtAl2013,GahlawatZhaoEtAl2020} can be used and updated online in the adaptive control loop. However, \emph{fast} closed-loop adaptive control with complex function approximators is hindered by the computational effort required to train them; this issue is exacerbated by the practical need for controller gain tuning. In our paper, we focus on offline meta-training of neural network features and controller gains from collected data, with well-known controller structures that can operate in fast closed-loops.

\subsection{Meta-Learning}
Meta-learning is the tool we use to inject the downstream adaptive control application into offline learning from data. Informally, meta-learning or ``learning to learn'' improves knowledge of how to best optimize a given meta-objective across \emph{different} tasks. In the literature, meta-learning has been formalized in various manners; we refer readers to \citet{HospedalesAntoniouEtAl2020} for a survey of them. In general, the algorithm chosen to solve a specific task is the \emph{base-learner}, while the algorithm used to optimize the meta-objective is the \emph{meta-learner}. In our work, when trying to make a dynamical system track several reference trajectories, each trajectory is associated with a ``task'', the adaptive tracking controller is the base-learner, and the average tracking error across all of these trajectories is the meta-objective we want to minimize.

Many works try to meta-learn a dynamics model offline that can best fit new input-output data gathered during a particular task. That is, the base- and meta-learners are \emph{regression-oriented}. \citet{BertinettoHenriquesEtAl2019} and \citet{LeeMajiEtAl2019} back-propagate through closed-form ridge regression solutions for few-shot learning, with a maximum likelihood meta-objective. \citet{OConnellShiEtAl2021} apply this same method to learn neural network features for nonlinear mechanical systems. \citet{HarrisonSharmaEtAl2018,HarrisonSharmaEtAl2018b} more generally back-propagate through a Bayesian regression solution to train a Bayesian prior dynamics model with nonlinear features. \citet{NagabandiClaveraEtAl2019} use a maximum likelihood meta-objective, and gradient descent on a multi-step likelihood objective as the base-learner. \citet{BelkhaleLiEtAl2021} also use a maximum likelihood meta-objective, albeit with the base-learner as a maximization of the Evidence Lower BOund (ELBO) over parameterized, task-specific variational posteriors; at test time, they perform latent variable inference online in a slow control loop.

\citet{FinnAbbeelEtAl2017,RajeswaranGhotraEtAl2017}, and \citet{ClaveraRothfussEtAl2018} meta-train a policy with the expected accumulated reward as the meta-objective, and a policy gradient step as the base-learner. These works are similar to ours in that they infuse offline learning with a more control-oriented flavour. However, while policy gradient methods are amenable to purely data-driven models, they beget slow control-loops due to the sampling and gradients required for each update. Instead, we back-propagate gradients through \emph{offline} closed-loop simulations to train adaptive controllers with well-known designs meant for fast online implementation. This yields a meta-trained adaptive controller that enjoys the performance of principled design inspired by the rich body of control-theoretical literature.

\section{Problem Statement}\label{sec:problem-statement}

In this paper, we are interested in controlling the continuous-time, nonlinear dynamical system
\begin{equation}\label{eq:nonlinear-system}
    \dot{x} = f(x, u, w),
\end{equation}
where $x(t) \in \R^n$ is the state, $u(t) \in \R^m$ is the control input, and $w(t) \in \R^d$ is some unknown disturbance. Specifically, for a given reference trajectory $r(t) \in \R^n$, we want to choose~$u(t)$ such that~$x(t)$ converges to~$r(t)$; we then say $u(t)$ makes the system~\cref{eq:nonlinear-system} track $r(t)$. 

Since $w(t)$ is unknown and possibly time-varying, we want to design a feedback law $u = \pi(x,r,a)$ with parameters $a(t)$ that are updated \emph{online} according to a chosen \emph{adaptation law} $\dot{a} = \rho(x,r,a)$. We refer to $(\pi,\rho)$ together as an \emph{adaptive controller}. For example, consider the control-affine system
\begin{equation}\label{eq:matched-uncertainty}
    \dot{x} = f_0(x) + B(x)(u + Y(x)w),
\end{equation}
where $f_0$, $B$, and $Y$ are known, possibly nonlinear maps. A reasonable feedback law choice would be
\begin{equation}\label{eq:adaptive-controller}
    u = \pi_0(x,r) - Y(x)a,
\end{equation}
where $\pi_0$ ensures $\dot{x} = f_0(x) + B(x)\pi_0(x,r)$ tracks $r(t)$, and the term $Y(x)a$ is meant to cancel $Y(x)w$ in~\cref{eq:matched-uncertainty}. For this reason, $Y(x)w$ is termed a matched uncertainty in the literature. If the adaptation law $\dot{a} = \rho(x,r,a)$ is designed such that~$Y(x(t))a(t)$ converges to~$Y(x(t))w(t)$, then we can use~\cref{eq:adaptive-controller} to make~\cref{eq:matched-uncertainty} track~$r(t)$. Critically, this is \emph{not} the same as requiring~$a(t)$ to converge to~$w(t)$. Since~$Y(x)w$ depends on~$x(t)$ and hence indirectly on the target~$r(t)$, the roles of feedback and adaptation are inextricably linked by the tracking control objective. Overall, learning in adaptive control is done on a ``need-to-know'' basis to cancel~$Y(x)w$ in \emph{closed-loop}, rather than to estimate~$w$ in open-loop.

\section{Bi-Level Meta-Learning}\label{sec:meta-learning}

We now describe some preliminaries on meta-learning akin to \citet{FinnAbbeelEtAl2017} and \citet{RajeswaranFinnEtAl2019}, so that we can apply these ideas in the next section to the adaptive control problem~\cref{eq:nonlinear-system} and in \cref{sec:baselines} to our baselines.

In machine learning, we typically seek some optimal parameters ${\varphi^* \in \argmin_\varphi \ell(\varphi,\mathcal{D})}$, where~$\ell$ is a scalar-valued loss function and~$\mathcal{D}$ is some data set. In meta-learning, we instead have a collection of loss functions~$\{\ell_i\}_{i=1}^M$, training data sets~$\{\mathcal{D}^{\mathrm{train}}_i\}_{i=1}^M$, and evaluation data sets $\{\mathcal{D}^{\mathrm{eval}}_i\}_{i=1}^M$, where each~$i$ corresponds to a task. Moreover, during each task~$i$, we can apply an adaptation mechanism ${\operatorname{\mathrm{Adapt}} : (\theta, \mathcal{D}^{\mathrm{train}}_i) \mapsto \varphi_i}$ to map so-called meta-parameters~$\theta$ and the task-specific training data~$\mathcal{D}^{\mathrm{train}}_i$ to task-specific parameters~$\varphi_i$. The crux of meta-learning is to solve the bi-level problem
\begin{equation}\label{eq:meta-learning}
\begin{aligned}
    \theta^* \in\, &\argmin_\theta \frac{1}{M}\rbr*{
        \sum_{i=1}^M \ell_i(\varphi_i, \mathcal{D}_i^{\mathrm{eval}})
        + \mu_\mathrm{meta}\norm{\theta}_2^2   
    }
    \\&\ \mathrm{s.t.}\enspace
    \varphi_i = \operatorname{\mathrm{Adapt}}(\theta,\mathcal{D}^{\mathrm{train}}_i)
\end{aligned},
\end{equation}
with regularization coefficient $\mu_\mathrm{meta} \geq 0$, thereby producing meta-parameters~$\theta^*$ that are on average well-suited to being adapted for each task. This motivates the moniker ``learning to learn'' for meta-learning. The optimization~\cref{eq:meta-learning} is the meta-problem, while the average loss is the meta-loss. The adaptation mechanism $\mathrm{Adapt}$ is termed the \emph{base-learner}, while the algorithm used to solve~\cref{eq:meta-learning} is termed the \emph{meta-learner} \cite{HospedalesAntoniouEtAl2020}.

Generally, the meta-learner is chosen to be some gradient descent algorithm. Choosing a good base-learner is an open problem in meta-learning research. \citet{FinnAbbeelEtAl2017} propose using a gradient descent step as the base-learner, such that $\varphi_i = \theta - \eta\grad_\theta\ell_i(\theta,\mathcal{D}_i^\mathrm{train})$ in~\cref{eq:meta-learning} with some learning rate~${\eta > 0}$. This approach is general in that it can be applied to any differentiable task loss functions. \citet{BertinettoHenriquesEtAl2019} and \citet{LeeMajiEtAl2019} instead study when the base-learner can be expressed as a convex program with a differentiable closed-form solution. In particular, they consider ridge regression with the hypothesis $\hat{y} = A g(x; \theta)$, where~$A$ is a matrix and $g(x; \theta)$ is some vector of nonlinear features parameterized by~$\theta$. For the base-learner, they use
\begin{equation}
    \varphi_i = \argmin_A \sum_{(x,y) \in \mathcal{D}_i^\mathrm{train}} \norm{y - A g(x; \theta)}_2^2 + \mu_\mathrm{ridge}\norm{A}_F^2,
\end{equation}
with regularization coefficient $\mu_\mathrm{ridge} > 0$ for the Frobenius norm $\norm{A}_F^2$, which admits a differentiable, closed-form solution. Instead of adapting~$\theta$ to each task~$i$ with a single gradient step, this approach leverages the convexity of ridge regression tasks to minimize the task loss analytically.

\section{Adaptive Control as a Base-Learner}

We now present the key idea of our paper, which uses meta-learning concepts introduced in \cref{sec:meta-learning} to tackle the problem of learning to control~\cref{eq:nonlinear-system}. For the moment, we assume we can simulate the dynamics function~$f$ in~\cref{eq:nonlinear-system} offline and that we have~$M$ samples $\{w_j(t)\}_{j=1}^M$ for $t \in [0,T]$ in~\cref{eq:nonlinear-system}; we will eliminate these assumptions in~\cref{sec:model-ensemble}. 

\subsection{Meta-Learning from Feedback and Adaptation}
\label{sec:meta-adaptive-control}
In meta-learning vernacular, we treat a reference trajectory~$r_i(t) \in \R^n$ and disturbance signal $w_j(t) \in \R^d$ together over some time horizon~$T > 0$ as the training data ${\mathcal{D}_{ij}^\mathrm{train} = \{r_i(t),w_j(t)\}_{t \in [0,T]}}$ for task~$(i,j)$. We wish to learn the static parameters $\theta \defn (\theta_\pi,\theta_\rho)$ of an adaptive controller
\begin{equation}\begin{aligned}
    u &= \pi(x,r,a; \theta_\pi), \\
    \dot{a} &= \rho(x,r,a; \theta_\rho),
\end{aligned}\end{equation}
such that $(\pi,\rho)$ engenders good tracking of $r_i(t)$ for $t \in [0,T]$ subject to the disturbance~$w_j(t)$. Our adaptation mechanism is the forward-simulation of our closed-loop system, \ie in~\cref{eq:meta-learning} we have $\varphi_{ij} = \{x_{ij}(t),a_{ij}(t),u_{ij}(t)\}_{t\in[0,T]}$, where
\begin{equation}\label{eq:simulate}
\begin{aligned}
    x_{ij}(t) &= x_{ij}(0) + \int_0^T f(x_{ij}(t),u_{ij}(t),w_j(t))\,dt, \\
    a_{ij}(t) &= a_{ij}(0) + \int_0^T \rho(x_{ij}(t),u_{ij}(t),w_j(t); \theta_\rho)\,dt, \\
    u_{ij}(t) &= \pi(x_{ij}(t),r_i(t),a_{ij}(t); \theta_\pi),
\end{aligned}\end{equation}
which we can compute with one of many Ordinary Differential Equation (ODE) solvers. For simplicity, we always set $x_{ij}(0) = r_i(0)$ and $a_{ij}(0) = 0$. Our task loss is simply the average tracking error for the same reference-disturbance pair, \ie $\mathcal{D}_{ij}^\mathrm{eval} = \{r_i(t),w_j(t)\}_{t \in [0,T]}$ and
\begin{equation}\label{eq:tracking-loss}
    \ell_{ij}(\varphi_{ij},\mathcal{D}_{ij}^\mathrm{eval}) 
    = \frac{1}{T} \int_0^{T} \rbr*{%
        \norm{x_{ij}(t) - r_i(t)}_2^2 + \alpha\norm{u_{ij}(t)}_2^2
    }\,dt,
\end{equation}
where $\alpha \geq 0$ regularizes the control effort $\frac{1}{T}\int_0^T\norm{u_{ij}(t)}_2^2\,dt$. This loss is inspired by the Linear Quadratic Regulator (LQR) from optimal control, and can be generalized to weighted norms. Assume we construct~$N$ reference trajectories~$\{r_i(t)\}_{i=1}^N$ and sample~$M$ disturbance signals~$\{w_j(t)\}_{j=1}^M$, thereby creating~$NM$ tasks. Combining \cref{eq:simulate} and \cref{eq:tracking-loss} for all $(i,j)$ in the form of \cref{eq:meta-learning} then yields the meta-problem
\begin{equation}\label{eq:meta-adaptive-control}
\begin{aligned}
    \min_\theta \
    &\frac{1}{NMT} \rbr*{ \sum_{i=1}^N \sum_{j=1}^M \int_0^{T} c_{ij}(t) \,dt + \mu_\mathrm{meta}\norm{\theta}_2^2 }
    \\&\mathrm{s.t.}\enspace\begin{aligned}[t]
        c_{ij} &= \norm{x_{ij} - r_i}_2^2 + \alpha\norm{u_{ij}}_2^2 \\
        \dot{x}_{ij} &= f(x_{ij}, u_{ij}, w_j),\  
            x_{ij}(0) = r_i(0) \\
        u_{ij} &= \pi(x_{ij}, r_i, a_{ij}; \theta_\pi) \\
        \dot{a}_{ij} &= \rho(x_{ij}, r_i, a_{ij}; \theta_\rho),\
            a_{ij}(0) = 0
    \end{aligned}
\end{aligned}
\end{equation}
Solving~\cref{eq:meta-adaptive-control} would yield parameters~$\theta = (\theta_\pi,\theta_\rho)$ for the adaptive controller~$(\pi,\rho)$ such that it works well on average in closed-loop tracking of~$\{r_i(t)\}_{i=1}^N$ for the dynamics~$f$, subject to the disturbances~$\{w_j(t)\}_{j=1}^M$. To learn the meta-parameters~$\theta$, we can perform gradient descent on~\cref{eq:meta-adaptive-control}. This requires back-propagating through an ODE solver, which can be done either directly or via the adjoint state method after solving the ODE forward in time~\cite{PontryaginBoltyanskiiEtAl1962,ChenRubanovaEtAl2018,AnderssonGillisEtAl2019,MillardHeidenEtAl2020}. In addition, the learning problem \cref{eq:meta-adaptive-control} is \emph{semi-supervised}, in that~$\{w_j(t)\}_{j=1}^M$ are labelled samples and~$\{r_i(t)\}_{i=1}^N$ can be chosen freely. If there are some specific reference trajectories we want to track at test time, we can use them in the meta-problem~\cref{eq:meta-adaptive-control}. This is an advantage of designing the offline learning problem in the context of the downstream control objective.

\subsection{Model Ensembling as a Proxy for Feedback Offline}
\label{sec:model-ensemble}

In practice, we cannot simulate the true dynamics~$f$ or sample an actual disturbance trajectory~$w(t)$ offline. Instead, we can more reasonably assume we have past data collected with some other, possibly poorly tuned controller. In particular, we make the following assumptions:
\begin{itemize}
    \item We have access to trajectory data $\{\mathcal{T}_j\}_{j=1}^M$, such that
    \begin{equation}\label{eq:trajectory-data}
        \mathcal{T}_j = \cbr[\big]{\rbr[\big]{
            t^{(j)}_k, x^{(j)}_k, u^{(j)}_k, t^{(j)}_{k+1}, x^{(j)}_{k+1}
        }}_{k=0}^{N_j-1},
    \end{equation}
    where~$x^{(j)}_k \in \R^n$ and $u^{(j)}_k \in \R^m$ were the state and control input, respectively, at time $t^{(j)}_k$. Moreover, $u^{(j)}_k$ was applied in a zero-order hold over $[t^{(j)}_k, t^{(j)}_{k+1})$, \ie $u(t) = u(t^{(j)}_k)$ for all $t \in [t^{(j)}_k, t^{(j)}_{k+1})$ along each trajectory $\mathcal{T}_j$. 
    
    \item During the collection of trajectory data $\mathcal{T}_j$, the disturbance~$w(t)$ took on a fixed, unknown value~$w_j$.
\end{itemize}
The second point is inspired by both meta-learning literature, where it is usually assumed the training data can be segmented according to the latent task, and adaptive control literature, where it is usually assumed that any unknown parameters are constant or slowly time-varying. These assumptions can be generalized to any collection of measured time-state-control transition tuples that can be segmented according to some latent task; in \cref{eq:trajectory-data} we consider when such tuples can be grouped into trajectories, since this is a natural manner in which data is collected from dynamical systems.

Inspired by \citet{ClaveraRothfussEtAl2018}, since we cannot simulate the true dynamics~$f$ offline, we propose to first train a \emph{model ensemble} from the trajectory data~$\{\mathcal{T}_j\}_{j=1}^M$ to roughly capture the distribution of $f(\cdot,\cdot,w)$ over possible values of the disturbance~$w$. Specifically, we fit a model~$\hat{f}_j(x,u; \psi_j)$ with parameters~$\psi_j$ to each trajectory~$\mathcal{T}_j$, and use this as a proxy for~$f(x,u,w_j)$ in~\cref{eq:meta-adaptive-control}. The meta-problem~\cref{eq:meta-adaptive-control} is now
\begin{equation}\label{eq:meta-adaptive-control-ensemble}
\begin{aligned}
    \min_\theta \
    &\frac{1}{NMT} \rbr*{ \sum_{i=1}^N \sum_{j=1}^M \int_0^{T} c_{ij}(t) \,dt + \mu_\mathrm{meta}\norm{\theta}_2^2 }
    \\&\mathrm{s.t.}\enspace\begin{aligned}[t]
        c_{ij} &= \norm{x_{ij} - r_i}_2^2 + \alpha\norm{u_{ij}}_2^2 \\
        \dot{x}_{ij} &= \hat{f}_j(x_{ij}, u_{ij}; \psi_j),\  
            x_{ij}(0) = r_i(0) \\
        u_{ij} &= \pi(x_{ij}, r_i, a_{ij}; \theta_\pi) \\
        \dot{a}_{ij} &= \rho(x_{ij}, r_i, a_{ij}; \theta_\rho),\
            a_{ij}(0) = 0
    \end{aligned}
\end{aligned}
\end{equation}
This form is still semi-supervised, since each model~$\hat{f}_j$ is dependent on the trajectory data~$\mathcal{T}_j$, while~$\{r_i\}_{i=1}^N$ can be chosen freely. The collection~$\{\hat{f}_j\}_{j=1}^M$ is termed a model ensemble. Empirically, the use of model ensembles has been shown to improve robustness to model bias and train-test data shift of deep predictive models~\cite{LakshminarayananPritzelEtAl2017} and policies in reinforcement learning~\cite{RajeswaranGhotraEtAl2017,KurutachClaveraEtAl2018,ClaveraRothfussEtAl2018}. To train the parameters~$\psi_j$ of model~$\hat{f}_j$ on the trajectory data~$\mathcal{T}_j$, we do gradient descent on the one-step prediction problem
\begin{equation}\label{eq:ensemble-training}
\begin{aligned}
    \min_{\psi_j} \
    &\frac{1}{N_j}\rbr*{ \sum_{k=0}^{N_j-1} \norm[\big]{x_{k+1}^{(j)} - \hat{x}_{k+1}^{(j)}}_2^2 + \mu_\mathrm{ensem}\norm{\psi_j}_2^2 }
    \\[-0.5em]&\mathrm{s.t.}\enspace\begin{aligned}[t]
        \hat{x}_{k+1}^{(j)} &= x_k^{(j)} + \int_{t_k^{(j)}}^{t_{k+1}^{(j)}} \hat{f}_j(x(t), u_k^{(j)}; \psi_j) \,dt
    \end{aligned}
\end{aligned}
\end{equation}
where $\mu_\mathrm{ensem} > 0$ regularizes~$\psi_j$. Since we meta-train~$\theta$ in \cref{eq:ensemble-training} to be adaptable to every model in the ensemble, we only need to roughly characterize how the dynamics $f(\cdot,\cdot,w)$ vary with the disturbance~$w$, rather than do exact model fitting of $\hat{f}_j$ to $\mathcal{T}_j$. Thus, we approximate the integral in \cref{eq:ensemble-training} with a single step of a chosen ODE integration scheme and back-propagate through this step, rather than use a full pass of an ODE solver.

\subsection{Incorporating Prior Knowledge About Robot Dynamics}
So far our method has been agnostic to the choice of adaptive controller~$(\pi,\rho)$. However, if we have some prior knowledge of the dynamical system~\cref{eq:nonlinear-system}, we can use this to make a good choice of structure for~$(\pi,\rho)$. Specifically, we now consider the large class of \emph{Lagrangian dynamical systems}, which includes robots such as manipulator arms and drones. The state of such a system is $x \defn (q,\dot{q})$, where $q(t) \in \R^{n_q}$ is the vector of generalized coordinates completely describing the configuration of the system at time~$t \in \R$. The nonlinear dynamics of such systems are fully described by
\begin{equation}\label{eq:lagrange}
    H(q)\ddot{q} + C(q,\dot{q})\dot{q} + g(q) 
    = f_\mathrm{ext}(q,\dot{q}) + \tau(u),
\end{equation}
where $H(q)$ is the positive-definite inertia matrix, $C(q,\dot{q})$ is the Coriolis matrix, $g(q)$ is the potential force, $\tau(u)$ is the generalized input force, and $f_\mathrm{ext}(q,\dot{q})$ summarizes any other external generalized forces. The vector $C(q,\dot{q})\dot{q}$ is uniquely defined, and the matrix $C(q,\dot{q})$ can always be chosen such that $\dot{H}(q,\dot{q}) - 2C(q,\dot{q})$ is skew-symmetric~\cite{SlotineLi1991}. \citet{SlotineLi1987} studied adaptive control for~\cref{eq:lagrange} under the assumptions:
\begin{itemize}
    \item The system~\cref{eq:lagrange} is fully-actuated, \ie $u(t) \in \R^{n_q}$ and ${\tau : \R^{n_q} \to \R^{n_q}}$ is invertible.
    \item The dynamics in~\cref{eq:lagrange} are linearly parameterizable, \ie
    \begin{equation}
        H(q)\dot{v} + C(q,\dot{q})v + g(q) - f_\mathrm{ext}(q,\dot{q}) 
        = Y(q,\dot{q},v,\dot{v})a,
    \end{equation}
    for some known matrix~$Y(q,\dot{q},v,\dot{v}) \in \R^{n_q \x p}$, any vectors $q,\dot{q},v,\dot{v} \in \R^{n_q}$, and constant unknown parameters~$a \in \R^p$.
    \item The reference trajectory for $x \defn (q,\dot{q})$ is of the form ${r = (q_d,\dot{q}_d)}$, where~$q_d$ is twice-differentiable.
\end{itemize}
Under these assumptions, the adaptive controller
\begin{equation}\label{eq:slotine-li-controller}
\begin{aligned}
    \tilde{q} &\defn q - q_d,\enspace
    s \defn \dot{\tilde{q}} + \Lambda{\tilde{q}},\enspace
    v \defn \dot{q}_d - \Lambda{\tilde{q}},
    \\
    u &= \inv{\tau}\rbr{ Y(q,\dot{q},v,\dot{v})\hat{a} - Ks },
    \\
    \dot{\hat{a}} &= -\Gamma \tran{Y(q,\dot{q},v,\dot{v})} s,
\end{aligned}\end{equation}
ensures $x(t) = (q(t),\dot{q}(t))$ converges to $r(t) = (q_d(t),\dot{q}_d(t))$, where $(\Lambda, K, \Gamma)$ are chosen positive-definite gain matrices.

The adaptive controller~\cref{eq:adaptive-controller} requires the nonlinearities in the dynamics~\cref{eq:lagrange} to be known a priori. While~\citet{NiemeyerSlotine1991} showed these can be systematically derived for~$H(q)$, $C(q,\dot{q})$, and~$g(q)$, there exist many external forces~$f_\mathrm{ext}(q,\dot{q})$ of practical importance in robotics for which this is difficult to do, such as aerodynamic and contact forces. Thus, we consider the case when~$H(q)$, $C(q,\dot{q})$, and~$g(q)$ are known and~$f_\mathrm{ext}(q,\dot{q})$ is unknown. Moreover, we want to approximate~$f_\mathrm{ext}(q,\dot{q})$ with the neural network
\begin{equation}
    \hat{f}_\mathrm{ext}(q,\dot{q}; A, \theta_y) = Ay(q,\dot{q}; \theta_y),
\end{equation}
where $y(q,\dot{q}; \theta) \in \R^p$ consists of all the hidden layers of the network parameterized by~$\theta_y$, and~$A \in \R^{n_q \x p}$ is the output layer. Inspired by~\cref{eq:slotine-li-controller}, we consider the adaptive controller
\begin{equation}\label{eq:slotine-li-controller-parametric}
\begin{aligned}
    \tilde{q} &\defn q - q_d,\enspace
    s \defn \dot{\tilde{q}} + \Lambda{\tilde{q}},\enspace
    v \defn \dot{q}_d - \Lambda{\tilde{q}},
    \\
    u &= \inv{\tau}\rbr{ H(q)\dot{v}\!+\! C(q,\dot{q})v\!+\! g(q)\!-\! Ay(q,\dot{q}; \theta_y)\!-\! Ks },
    \\
    \dot{A} &= \Gamma s \tran{y(q,\dot{q}; \theta_y)},
\end{aligned}\end{equation}
If $f_\mathrm{ext}(q,\dot{q}) = \hat{f}_\mathrm{ext}(q,\dot{q}; A, \theta_y)$ for fixed values~$\theta_y$ and~$A$, then the adaptive controller~\cref{eq:slotine-li-controller-parametric} guarantees tracking convergence~\cite{SlotineLi1987}. In general, we do not know such a value for~$\theta_y$, and we must choose the gains~$(\Lambda, K, \Gamma)$. Since~\cref{eq:slotine-li-controller-parametric} is parameterized by $\theta \defn (\theta_y, \Lambda, K, \Gamma)$, we can train~\cref{eq:slotine-li-controller-parametric} with the method described in \crefrange{sec:meta-adaptive-control}{sec:model-ensemble}. While 
for simplicity we consider known $H(q)$, $C(q,\dot{q})$, and $g(q)$, we can extend to the case when they are linearly parameterizable, \eg $H(q)\dot{v} + C(q,\dot{q})v + g(q) = Y(q,\dot{q},v,\dot{v})a\ $ with  $Y(q,\dot{q},v,\dot{v})$ a matrix of known features systematically computed as by \citet{NiemeyerSlotine1991}. In this case, we would then maintain a separate adaptation law~${\dot{\hat{a}} = -P \tran{Y(q,\dot{q},v,\dot{v})} s}$ with adaptation gain $P \succ 0$ in our proposed adaptive controller \cref{eq:slotine-li-controller-parametric}.

\section{Experiments}\label{sec:experiments}

We evaluate our method in simulation on a Planar Fully-Actuated Rotorcraft (PFAR) with degrees of freedom ${q \defn (x,y,\phi)}$ governed by the nonlinear equations of motion
\begin{equation}\label{eq:pfar}
    \pmx{\ddot{x} \\ \ddot{y} \\ \ddot{\phi}} + g
    = \underbrace{\bmx{
        \cos\phi  & -\sin\phi   & 0 \\
        \sin\phi  &  \cos\phi   & 0 \\
        0         & 0           & 1
    }}_{\eqqcolon R(\phi)}u + f_\mathrm{ext},
\end{equation}
where $(x,y)$ is the position of the center of mass in the inertial frame, $\phi$ is the roll angle, $g = (0, 9.81, 0)~\mathrm{m/s^2}$ is the gravitational acceleration in vector form, $R(\phi)$ is a rotation matrix, $f_\mathrm{ext}$ is some unknown external force, and ${u = (u_1,u_2,u_3)}$ are the normalized thrust along the body $x$-axis, thrust along the body $y$-axis, and torque about the center of mass, respectively. We depict an exemplary PFAR design in \cref{fig:pfar} inspired by thriving interest in fully- and over-actuated multirotor vehicles in the robotics literature~\cite{RyllMuscioEtAl2017,KamelVerlingEtAl2018,BrescianiniDAndrea2018,ZhengTanEtAl2020,RashadGoerresEtAl2020}. The simplified system in \cref{eq:pfar} is a fully-actuated variant of the classic Planar Vertical Take-Off and Landing (PVTOL) vehicle~\cite{HauserSastryEtAl1992}. In our simulations, $f_\mathrm{ext}$ is a mass-normalized quadratic drag force, due to the velocity of the PFAR relative to wind blowing at a velocity~$w \in \R$ along the inertial $x$-axis. Specifically,
\begin{equation}\label{eq:drag}
\begin{aligned}
    v_1 &= (\dot{x} - w)\cos\phi + \dot{y}\sin\phi, \\
    v_2 &= -(\dot{x} - w)\sin\phi + \dot{y}\cos\phi, \\
    f_\mathrm{ext} &= -\bmx[0.8]{
        \cos\phi  & -\sin\phi \\
        \sin\phi  &  \cos\phi \\
        0         & 0
    }\pmx{\beta_1v_1\abs{v_1} \\ \beta_2v_2\abs{v_2}},
\end{aligned}\end{equation}
where $\beta_1, \beta_2 > 0$ are aggregate coefficients; we use ${\beta_1 = 0.1}$ and ${\beta_2 = 1}$ in all of our simulations.

\subsection{Baselines}\label{sec:baselines}

We compare our meta-trained adaptive controller in trajectory tracking tasks for~\cref{eq:pfar} against two baseline controllers:

\paragraph*{PID Control} Our simplest baseline is a Proportional-Integral-Derivative (PID) controller with feed-forward, \ie
\begin{equation}\label{eq:pid}
    u = \tran{R(\phi)}\rbr*{
        g + \ddot{q}_d - K_P\tilde{q} - K_I\!\int_0^t\tilde{q}(\xi)\,d\xi - K_D\dot{\tilde{q}}
    },
\end{equation}
with gains $K_P, K_I, K_D \succ 0$. For $f_\mathrm{ext}(q,\dot{q}) \equiv 0$, this controller feedback-linearizes the dynamics~\cref{eq:pfar} such that the error~$\tilde{q} \defn q - q_d$ is governed by an exponentially stable ODE. The integral term must compensate for $f_\mathrm{ext}(q,\dot{q}) \not\equiv 0$.

\paragraph*{Adaptive Control via Meta-Ridge Regression (ACMRR)} This baseline is a slightly modified\footnote{
    Unlike \citet{OConnellShiEtAl2021}, we do not assume access to direct measurements of the external force~$f_\mathrm{ext}$. Also, they use a more complex form of~\cref{eq:slotine-li-controller} with better \emph{parameter estimation} properties when the dynamics are linearly parameterizable with \emph{known} nonlinear features~\cite{SlotineLi1989}. While we could use a parametric form of this controller in place of~\cref{eq:slotine-li-controller-parametric}, we forgo this in favour of a simpler presentation, since we focus on offline \emph{control-oriented} meta-learning of \emph{approximate} features.
} version of the approach taken by \citet{OConnellShiEtAl2021}, which applies the work on using ridge regression as a base-learner from~\citet{BertinettoHenriquesEtAl2019,LeeMajiEtAl2019} and \citet{HarrisonSharmaEtAl2018} to learn the parametric features~$y(q,\dot{q};\theta_y)$. Specifically, for a given trajectory~$\mathcal{T}_j$ and these features, the last layer~$A$ is specified as the best ridge regression fit to some subset of points in~$\mathcal{T}_j$. The features~$y(q,\dot{q};\theta_y)$ are then used in the adaptive controller~\cref{eq:slotine-li-controller-parametric}.

To clarify, we describe ACMRR with the meta-learning language from~\cref{sec:meta-learning}, specifically for the PFAR dynamics in~\cref{eq:pfar}. Let~${\mathcal{K}_j^\mathrm{ridge} \subset \{k\}_{k=0}^{\abs{\mathcal{T}_j}-1}}$ denote the indices of transition tuples in some subset of $\mathcal{T}_j$. The Euler approximation
\begin{equation}\label{eq:euler}
    \hat{\dot{q}}_{k+1}^{(j)}(A)
    \defn \dot{q}_k^{(j)}{+}\Delta t_k^{(j)}\!\rbr[\big]{
            R(\phi_k^{(j)}) u_k^{(j)}{+}Ay(q_k^{(j)}\!\!,\dot{q}_k^{(j)};\theta_y)
    }
\end{equation}
with $\Delta t_k^{(j)} \defn t_{k+1}^{(j)} - t_k^{(j)}$ is used in the task loss
\begin{equation}\label{eq:lstsq-task-loss}
    \ell_j(A_j, \mathcal{T}_j)
    = \frac{1}{\abs{\mathcal{T}_j}}\sum_{k = 0}^{\abs{\mathcal{T}_j} - 1} \norm{\dot{q}_{k+1}^{(j)} - \dot{\hat{q}}_{k+1}^{(j)}(A_j)}_2^2
\end{equation}
alongside the adaptation mechanism
\begin{equation}\label{eq:lstsq-adapt}
    A_j = \argmin_A \! \sum_{k \in \mathcal{K}_j^\mathrm{ridge}} \norm{
        \dot{q}_{k+1}^{(j)} - \hat{\dot{q}}_{k+1}^{(j)}(A)
    }_2^2 + \mu_\mathrm{ridge}\norm{A}_F^2.
\end{equation}
The adaptation mechanism~\cref{eq:lstsq-adapt} can be solved and differentiated in closed-form for any $\mu_\mathrm{ridge} > 0$ via the normal equations, since~$\hat{\dot{q}}_{k+1}^{(j)}(A)$ is linear in~$A$; indeed, only \emph{linear} integration schemes can be substituted into~\cref{eq:euler}. The meta-problem for ACMRR takes the form of~\cref{eq:meta-learning} with meta-parameters~$\theta_y$ for the features~$y(q,\dot{q};\theta_y)$, the task loss~\cref{eq:lstsq-task-loss}, and the adaptation mechanism~\cref{eq:lstsq-adapt}. The meta-parameters~$\theta_y$ are trained via gradient descent on this meta-problem, and then deployed online in the adaptive controller~\cref{eq:slotine-li-controller-parametric}.

ACMRR suffers from a \emph{fundamental mismatch} between its regression-oriented meta-problem and the online problem of adaptive trajectory tracking control. The ridge regression base-learner~\cref{eq:lstsq-adapt} suggests that~$A$ should best fit the input-output trajectory data in a regression sense. However, as we mentioned in~\cref{sec:problem-statement}, adaptive controllers such as the one in~\cref{eq:slotine-li-controller-parametric} learn on a ``need-to-know'' basis for the primary purpose of control rather than regression. As we discuss in \cref{sec:results-discussion}, since our control-oriented approach uses a meta-objective indicative of the downstream closed-loop tracking control objective, we achieve better tracking performance than this baseline at test time.

\subsection{Data Generation and Training}

To train the meta-parameters $\theta_\mathrm{ours} \defn (\theta_y, \Lambda, K, \Gamma)$ in our method, and the meta-parameters $\theta_\mathrm{ACMRR} \defn \theta_y$ in the ACMRR baseline, we require trajectory data $\{\mathcal{T}_j\}_{j=1}^M$ of the form \cref{eq:trajectory-data}. To this end, we synthesize~$\mathcal{T}_j$ as follows:
\begin{enumerate}
    \item Generate a uniform random walk of six $(x,y,\phi)$ points.
    
    \item Fit a 30-second, smooth, polynomial spline trajectory~$q_d(t) \in \R^3$ to the random walk with minimum snap in~$(x,y)$ and minimum acceleration in~$\phi$, according to the work by \citet{MellingerKumar2011} and \citet{RichterBryEtAl2013}.
    
    \item Sample a wind velocity $w \in \R$ from the training distribution in~\cref{fig:wind_dist}, and simulate the dynamics~\cref{eq:pfar} with the external force~\cref{eq:drag} and the Proportional-Derivative (PD) tracking controller
    \begin{equation}
        u = \tran{R(\phi)}\rbr*{g - k_P\tilde{q} - k_D\dot{\tilde{q}}}
    \end{equation}
    in a zero-order hold at $100~\mathrm{Hz}$, with ${k_P = 10}$ and ${k_D = 0.1}$. This controller represents a ``first try'' at controlling the system~\cref{eq:pfar} in order to collect data. We record time, state, and control input measurements at $100~\mathrm{Hz}$. 
\end{enumerate}
We record $500$~such trajectories and then sample~$M$ of them to form the training data~$\{\mathcal{T}_j\}_{j=1}^M$ for various~$M$ to evaluate the sample efficiency of our method and the ACMRR baseline. We present hyperparameter choices and training details for each method in~\cref{app:training}. We highlight here that:
\begin{itemize}
    \item To train the positive-definite gains~$(\Lambda,K,\Gamma)$ via gradient-based optimization in our method, we use an unconstrained log-Cholesky parameterization for each gain\footnote{
        Any $n \times n$ positive-definite matrix~$Q$ can be uniquely defined by $\frac{1}{2}n(n+1)$ parameters. For $n=2$, the log-Cholesky parameterization of $Q$ is $Q = L\tran{L}$ with $L = \bmx{\exp(\theta_1) & 0 \\ \theta_2 & \exp(\theta_3)}$ and unconstrained parameters $\theta \in \R^3$~\cite{PinheiroBates1996}.
    }.
    \item To compute the integral in the meta-problem~\cref{eq:meta-adaptive-control-ensemble} for our method, we use a fourth-order Runge-Kutta scheme with a fixed time step of $0.01~\mathrm{s}$. We back-propagate gradients through this computation in a manner similar to~\citet{ZhuangDvornekEtAl2020}, rather than using the adjoint method for neural ODEs~\cite{ChenRubanovaEtAl2018}, due to our observation that the backward pass is sensitive to any numerical error accumulated along the forward pass during closed-loop control simulations.
\end{itemize}

\subsection{Testing with Distribution Shift}\label{sec:testing}
\begin{figure}
    \centering
    \includegraphics[width=\columnwidth]{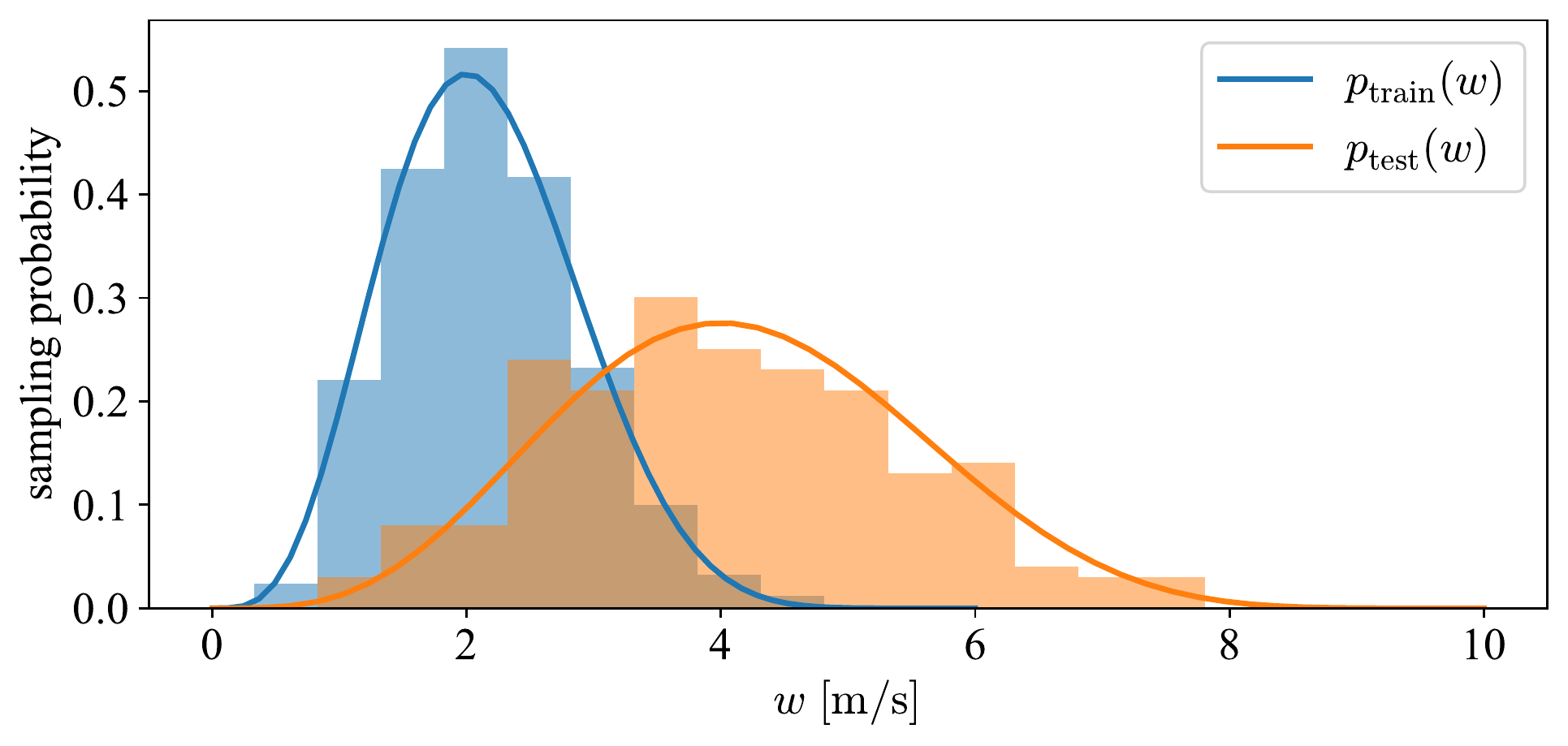}
    \vspace{-2em}
    \caption{Training distribution~$p_\mathrm{train}$ and test distribution~$p_\mathrm{test}$ for the wind velocity~$w$ along the inertial $x$-axis. Both are scaled beta distributions; $p_\mathrm{train}$ is supported on the interval $[0, 6]~\mathrm{m/s}$ with shape parameters~$(\alpha,\beta) = (5,9)$, while $p_\mathrm{test}$ is supported on the interval $[0, 10]~\mathrm{m/s}$ with shape parameters~$(\alpha,\beta) = (5,7)$. The normalized histograms show the distribution of the actual wind velocity samples in the training and test data for a single seed, and highlight the out-of-distribution samples (\ie relative to $p_\mathrm{train}$) that occur during testing.}
    \label{fig:wind_dist}
\end{figure}
For testing, we simulate tracking of~$200$ smooth, $10$-second reference trajectories \emph{different} from those in the training data, using our meta-parameters~$\theta_\mathrm{ours} \defn (\theta_y, \Lambda, K, \Gamma)$ with the adaptive controller~\cref{eq:slotine-li-controller-parametric}, the ACMRR meta-parameters ${\theta_\mathrm{ACMRR} \defn \theta_y}$ with the adaptive controller~\cref{eq:slotine-li-controller-parametric}, and the PID controller~\cref{eq:pid}; each controller is implemented at~$100~\mathrm{Hz}$. As shown in \cref{fig:wind_dist}, we sample wind velocities at test time from a distribution \emph{different} from that used for the training data~$\{\mathcal{T}_j\}_{j=1}^M$; in particular, the test distribution has a higher mode and larger support than the training distribution, thereby producing so-called out-of-distribution wind velocities at test time. This ensures the generalizability of each approach is tested, which is a particularly important concept in meta-learning literature~\cite{HospedalesAntoniouEtAl2020}.

A strength of our method is that it meta-learns both features~$y(q,\dot{q};\theta_y)$ and gains~$(\Lambda,K,\Gamma)$ offline, while both the PID and ACMRR baselines require gain tuning in practice by interacting with the real system. However, for the sake of comparison, we test every method with various manually chosen gains and our method with our meta-learned gains, on the same set of test trajectories. In particular, for the PID controller~\cref{eq:pid}, we set ${K_P = K\Lambda + \Gamma}$, ${K_I = \Gamma\Lambda}$, and ${K_D = K + \Lambda}$, which makes it equivalent to the adaptive controller~\cref{eq:slotine-li-controller-parametric} for the dynamics~\cref{eq:pfar} with $y(q,\dot{q};\theta_y) \equiv 1$ (\ie constant features), $\tilde{q}(0) = 0$ (\ie zero initial position error), and $A(0) = 0$ (\ie adapted parameters are initially zero). We always set initial conditions for the system such that~$\tilde{q}(0) = \dot{\tilde{q}}(0) = 0$, and $A(0) = 0$.

\subsection{Results and Discussion}\label{sec:results-discussion}

\begin{figure}
    \centering
    \includegraphics[width=\columnwidth]{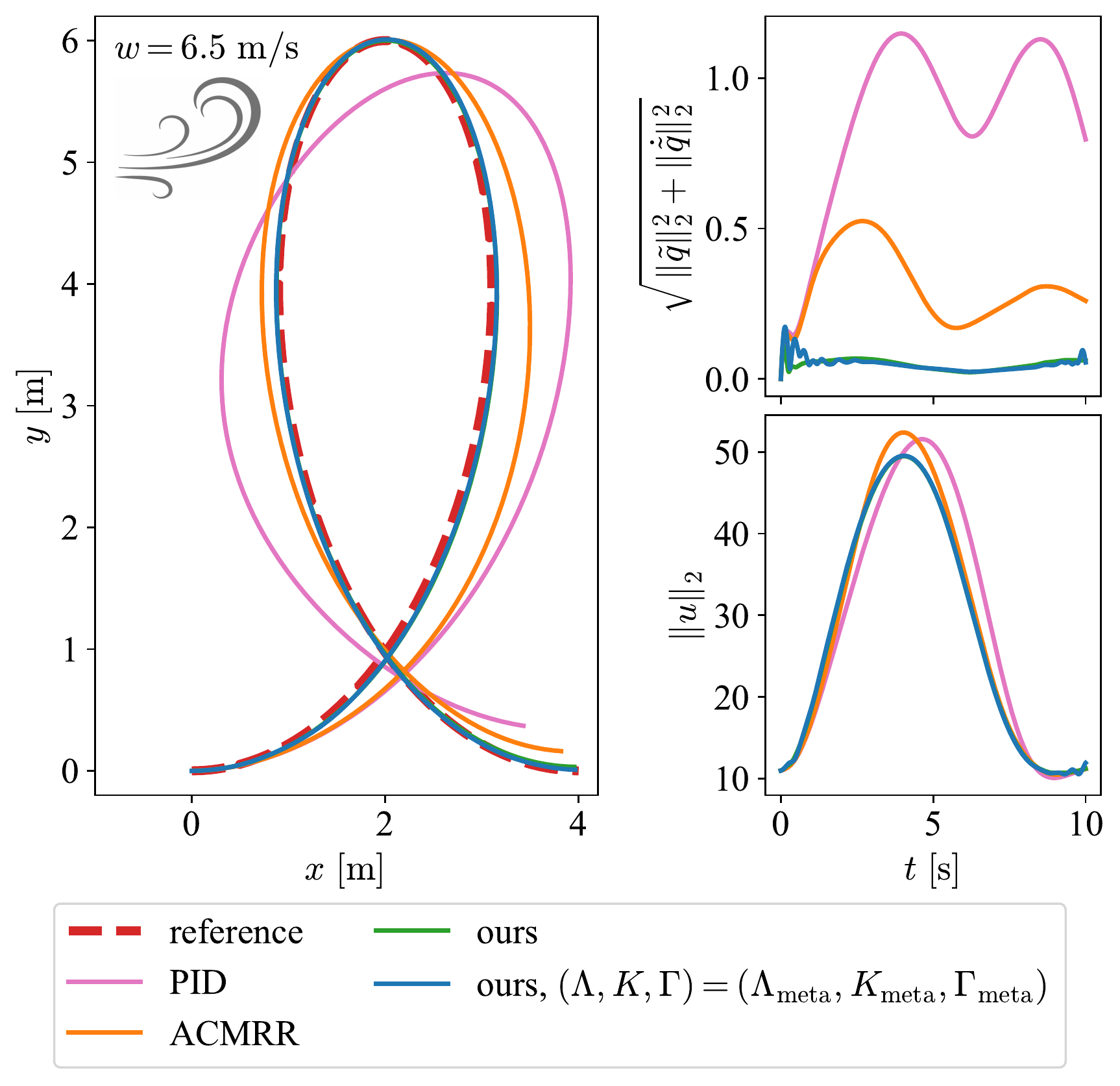}
    \vspace{-2em}
    \caption{%
        Tracking results for the PFAR on a test trajectory with ${w = 6.5~\mathrm{m/s}}$, $M=10$, and $(\Lambda, K, \Gamma) = (I, 10I, 10I)$. We also apply our meta-learned features and our meta-learned gains $(\Lambda_\mathrm{meta}, K_\mathrm{meta}, \Gamma_\mathrm{meta})$. All of the methods expend similar control efforts. However, with our meta-learned features, the tracking error ${\norm{x - r}_2 = \sqrt{\norm{\tilde{q}}_2^2 + \norm{\dot{\tilde{q}}}_2^2}}$ (where $x$ is overloaded to denote both position $x \in \R$ and state $x = (q,\dot{q})$) quickly decays after a short transient, while the effect of the wind pushing the vehicle to the right is more pronounced for the baselines.
    }\label{fig:results_traj}
    \vspace{-0.4em}
\end{figure}

\begin{figure*}
    \centering
    \includegraphics[width=\textwidth]{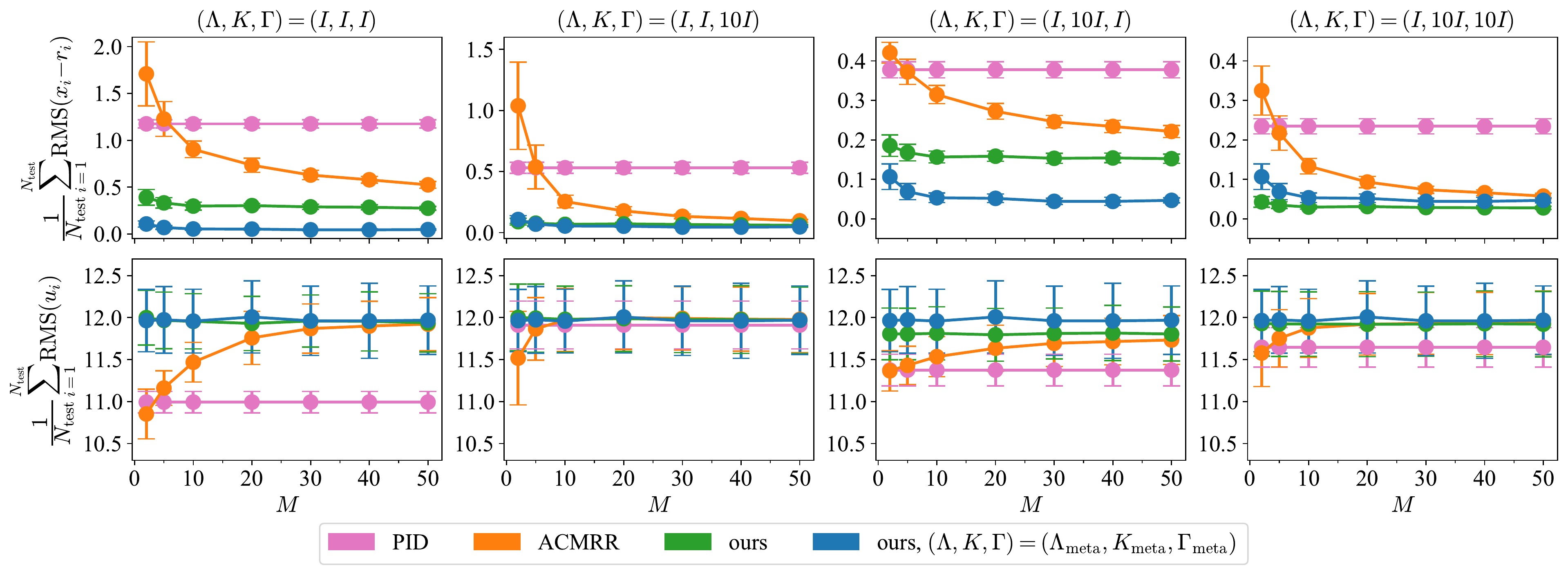}
    \vspace{-2em}
    \caption{%
        Line plots of the average RMS tracking error $\frac{1}{N_\mathrm{test}}\sum_{i=1}^{N_\mathrm{test}}\mathrm{RMS}(x_i{-}r_i)$ and average control effort  $\frac{1}{N_\mathrm{test}}\sum_{i=1}^{N_\mathrm{test}}\mathrm{RMS}(u_i)$ across $N_\mathrm{test}=200$ test trajectories versus the number of trajectories~$M \in \{2, 5, 10, 20, 30, 40, 50\}$ in the training data. For each method, we try out various control gains $(\Lambda, K, \Gamma)$. With our method, we also use our meta-learned gains $(\Lambda_\mathrm{meta}, K_\mathrm{meta}, \Gamma_\mathrm{meta})$. The results for the PID controller do not vary with~$M$ since it does not require any training. Dots and error bars denote means and standard deviations, respectively, across 10~random seeds. Our experiments are done in Python using NumPy~\cite{HarrisMillmanEtAl2020} and JAX~\cite{BradburyFrostigEtAl2018}. We use the explicit nature of Pseudo-Random Number Generation (PRNG) in JAX to set a seed prior to training, and then methodically branch off the associated PRNG key as required throughout the train-test experiment pipeline. Thus, all of our results can be easily reproduced; code to do so is provided at~\url{https://github.com/StanfordASL/Adaptive-Control-Oriented-Meta-Learning}.
    }\label{fig:results}
    \vspace{-1em}
\end{figure*}

We first provide a qualitative plot of tracking results for each method on a single test trajectory in \cref{fig:results_traj}, which clearly shows that our meta-learned features induce better tracking results than the baselines, while requiring a similar expenditure of control effort. For our method, the initial transient decays quickly, thereby demonstrating \emph{fast} adaptation and the potential to handle even time-varying disturbances.

For a more thorough analysis, we consider the Root-Mean-Squared (RMS) tracking error and control effort for each test trajectory~$\{r_i\}_{i=1}^{N_\mathrm{test}}$; for any vector-valued signal~$h(t)$ and sampling times~$\{t_k\}_{k=0}^N$, we define
\begin{equation}
    \mathrm{RMS}(h) \defn \sqrt{ \frac{1}{N} \sum_{k = 0}^N \norm{h(t_k)}_2^2 }.
\end{equation}
We are interested in $\mathrm{RMS}(x_i{-}r_i)$ and $\mathrm{RMS}(u_i)$, where $x_i(t) = (q_i(t),\dot{q}_i(t))$ and $u_i(t)$ are the resultant state and control trajectories from tracking ${r_i(t) = (q_{d,i}(t),\dot{q}_{d,i}(t))}$. In \cref{fig:results}, we plot the averages $\frac{1}{N_\mathrm{test}}\sum_{i=1}^{N_\mathrm{test}}\mathrm{RMS}(x_i{-}r_i)$ and $\frac{1}{N_\mathrm{test}}\sum_{i=1}^{N_\mathrm{test}}\mathrm{RMS}(u_i)$ across $N_\mathrm{test}=200$ test trajectories for each method. For our method and the ACMRR baseline, we vary the number of training trajectories~$M$, and thus the number of wind velocities from the training distribution in~\cref{fig:wind_dist} implicitly present in the training data. From~\cref{fig:results}, we observe the PID controller generally yields the highest tracking error, thereby indicating the utility of meta-learning features~$y(q,\dot{q};\theta_y)$ to better compensate for~$f_\mathrm{ext}(q,\dot{q})$. We further observe in~\cref{fig:results} that, regardless of the control gains, using our features~$y(q,\dot{q};\theta_y)$ in the adaptive controller~\cref{eq:slotine-li-controller-parametric} yields the lowest tracking error. Moreover, using our features with our meta-learned gains yields the lowest tracking error in all but one case. Our features induce a slightly higher control effort with a greater standard deviation across random seeds, especially when used with our meta-learned gains~$(\Lambda_\mathrm{meta}, K_\mathrm{meta}, \Gamma_\mathrm{meta})$. This is most likely since the controller can better match the disturbance~$f_\mathrm{ext}(q,\dot{q})$ with our features, and is an acceptable trade-off given that our primary objective is trajectory tracking. In addition, when using our features with manually chosen or our meta-learned gains, the tracking error remains relatively constant over~$M$; conversely, the tracking error for the ACMRR baseline is higher for lower~$M$, and only reaches the performance of our method with certain gains for large~$M$. Overall, our results indicate:
\begin{itemize}
    \item The features~$y(q,\dot{q};\theta_y)$ meta-learned by our control-oriented method are \emph{better conditioned for closed-loop tracking control} across a range of controller gains, particularly in the face of a distributional shift between training and test scenarios.
    \item The gains meta-learned by our method are competitive \emph{without manual tuning}, and thus can be deployed immediately or serve as a good initialization for further fine-tuning.
    \item Our control-oriented meta-learning method is \emph{sample-efficient} with respect to how much system variability is implicitly captured in the training data.
\end{itemize}
We again stress these comparisons could only be done by tuning the control gains for the baselines, which in practice would require interaction with the real system and hence further data collection. Thus, the fact that our control-oriented method can meta-learn good control gains \emph{offline} is a key advantage over regression-oriented meta-learning.

\section{Conclusions \& Future Work}\label{sec:conclusion}

In this work, we formalized control-oriented meta-learning of adaptive controllers for nonlinear dynamical systems, offline from trajectory data. The procedure we presented is general and uses adaptive control as the base-learner to attune learning to the downstream control objective. We then specialized our procedure to the class of nonlinear \emph{mechanical} systems, with a well-known adaptive controller parameterized by control gains and nonlinear model features. We demonstrated that our control-oriented meta-learning method engenders better closed-loop tracking control performance at test time than when learning is done for the purpose of model regression.

There are a number of exciting future directions for this work. In particular, we are interested in control-oriented meta-learning with \emph{constraints}, such as for adaptive Model Predictive Control (MPC) \cite{AdetolaGuay2011,BujarbaruahZhangEtAl2018,SolopertoKohlerEtAl2019,KohlerKottingEtAl2020,SinhaHarrisonEtAl2021} with state and input constraints. Back-propagating through such a controller would leverage recent work on differentiable convex optimization~\cite{AmosRodriguezEtAl2018,AgrawalBarrattEtAl2019,AgrawalBarrattEtAl2020}. We could also back-propagate through parameter constraints; for example, physical consistency of adapted inertial parameters can be enforced as Linear Matrix Inequality (LMI) constraints that reduce overfitting and improve parameter convergence \cite{WensingKimEtAl2017}. In addition, we want to extend our meta-learning approach to adaptive control for underactuated systems. Underactuation is a fundamental challenge for adaptive control since model uncertainties must satisfy certain matching conditions so that they can be cancelled stably by the controller~\cite{LopezSlotine2021,SinhaHarrisonEtAl2021}. To this end, we want to explore how meta-learning can be used to learn adaptive controllers defined in part by parametric stability and stabilizability certificates, such as Lyapunov functions and Control Contraction Metrics (CCMs)~\cite{ManchesterSlotine2017}. This could build off of existing work on learning such certificates from data~\cite{RichardsBerkenkampEtAl2018,SinghRichardsEtAl2020,BoffiTuEtAl2020,TsukamotoChungEtAl2021}.

\vfill

\section*{Acknowledgements}
We thank Masha Itkina for her invaluable feedback, and Matteo Zallio for his expertise in crafting~\cref{fig:pfar}. This research was supported in part by the National Science Foundation (NSF) via Cyber-Physical Systems (CPS) award \#1931815 and Energy, Power, Control, and Networks  (EPCN) award \#1809314, and the National Aeronautics and Space Administration (NASA) University Leadership Initiative via grant \#80NSSC20M0163. Spencer M.~Richards was also supported in part by the Natural Sciences and Engineering Research Council of Canada (NSERC). This article solely reflects our own opinions and conclusions, and not those of any NSF, NASA, or NSERC entity.

\bibliographystyle{plainnat}
\bibliography{ASL_papers,main}

\appendices
\crefalias{section}{appendix}

\section{Training Details}\label{app:training}

\paragraph*{Our Method} To meta-train $\theta_\mathrm{ours} \defn (\theta_y, \Lambda, K, \Gamma)$, we first train an ensemble of~$M$ models~$\{\hat{f}_j\}_{j=1}^M$, one for each trajectory $\mathcal{T}_j$, via gradient descent on the regression objective~\cref{eq:ensemble-training} with $\mu_\mathrm{ensem} = 10^{-4}$ and a single fourth-order Runge-Kutta step to approximate the integral over $t^{(j)}_{k+1} - t^{(j)}_k = 0.01~\mathrm{s}$. We set each~$\hat{f}_j$ as a feed-forward neural network with~$2$ hidden layers, each containing~$32$ $\tanh(\cdot)$ neurons. We perform a random $75\%/25\%$ split of the transition tuples in $\mathcal{T}_j$ into a training set~$\mathcal{T}_j^\mathrm{train}$ and validation set~$\mathcal{T}_j^\mathrm{valid}$, respectively. We do batch gradient descent via Adam~\cite{KingmaBa2015} on~$\mathcal{T}_j^\mathrm{train}$ with a step size of~$10^{-2}$, over $1000$~epochs with a batch size of $\floor{0.25\abs{\mathcal{D}_j^\mathrm{train}}}$, while recording the regression loss with~$\mu_\mathrm{ensem} = 0$ on~$\mathcal{T}_j^\mathrm{valid}$. We proceed with the parameters for~$\hat{f}_j$ corresponding to the lowest recorded validation loss.

With the trained ensemble~$\{\hat{f}_j\}_{j=1}^M$, we can now meta-train $\theta_\mathrm{ours} \defn (\theta_y, \Lambda, K, \Gamma)$. First, we randomly generate ${N = 10}$ smooth reference trajectories~$\{r_i\}_{i=1}^N$ in the same manner as above, each with a duration of~$T = 5~\mathrm{s}$. We then randomly sub-sample $N_\mathrm{train} = \floor{0.75 N} = 7$ reference trajectories and $M_\mathrm{train} = \floor{0.75 M}$ models from the ensemble to form the meta-training set $\{(r_i,\hat{f}_j)\}_{i=1,j=1}^{N_\mathrm{train}, M_\mathrm{train}}$, while the remaining models and reference trajectories form the meta-validation set. We set $Ay(q,\dot{q};\theta_y)$ as a feed-forward neural network with~$2$ hidden layers of~$32$ $\tanh(\cdot)$ neurons each, where the adapted parameters~$A(t) \in \R^{n_q \x 32}$ with $n_q = 3$ serve as the output layer. We use an unconstrained log-Cholesky parameterization for each of the positive-definite gains~$(\Lambda, K, \Gamma)$. We set up the meta-problem~\cref{eq:meta-adaptive-control-ensemble} using $\{(r_i,\hat{f}_j)\}_{i=1,j=1}^{N_\mathrm{train}, M_\mathrm{train}}$, $\alpha = 10^{-3}$, $\mu_\mathrm{meta} = 10^{-4}$, and the adaptive controller~\cref{eq:slotine-li-controller-parametric}. We then perform gradient descent via Adam with a step size of~$10^{-2}$ to train~$\theta_\mathrm{ours} \defn (\theta_y, \Lambda, K, \Gamma)$. We compute the integral in~\cref{eq:meta-adaptive-control-ensemble} via a fourth-order Runge-Kutta integration scheme with a fixed time step of $0.01~\mathrm{s}$. We back-propagate gradients through this computation in a manner similar to~\citet{ZhuangDvornekEtAl2020}, rather than using the adjoint method for neural ODEs~\cite{ChenRubanovaEtAl2018}, due to our observation that the backward pass is sensitive to any numerical error accumulated along the forward pass during closed-loop control simulations. We perform~$500$ gradient steps while recording the meta-loss from~\cref{eq:meta-adaptive-control-ensemble} with~$\mu_\mathrm{meta} = 0$ on the meta-validation set, and take the best meta-parameters~$\theta_\mathrm{ours}$ as those corresponding to the lowest recorded validation loss.

\paragraph*{ACMRR Baseline} To meta-train $\theta_\mathrm{ACMRR} \defn \theta_y$, we first perform a random $75\%/25\%$ split of the transition tuples in each trajectory $\mathcal{T}_j$ to form a meta-training set $\{\mathcal{T}_j^\mathrm{meta\text{-}train}\}_{j=1}^M$ and a meta-validation set $\{\mathcal{T}_j^\mathrm{meta\text{-}valid}\}_{j=1}^M$. We set $Ay(q,\dot{q};\theta_y)$ as a feed-forward neural network with~$2$ hidden layers of~$32$ $\tanh(\cdot)$ neurons each, where the adapted parameters~$A(t) \in \R^{n_q \x 32}$ with $n_q = 3$ serve as the output layer. We construct the meta-problem~\cref{eq:meta-learning} using the task loss~\cref{eq:lstsq-task-loss}, the adaptation mechanism~\cref{eq:lstsq-adapt}, and $\mu_\mathrm{meta} = 10^{-4}$; for this, we use transition tuples from $\mathcal{T}_j^\mathrm{meta\text{-}train}$. We then meta-train~$\theta_\mathrm{ACMRR}$ via gradient descent using Adam with a step-size of~$10^{-2}$ for $5000$ steps; at each step, we randomly sample a subset of $\floor{0.25\abs{\mathcal{T}_j^\mathrm{meta\text{-}train}}}$ tuples from $\mathcal{T}_j^\mathrm{meta\text{-}train}$ to use in the closed-form ridge regression solution. We also record the meta-loss with $\mu_\mathrm{meta} = 0$ on the meta-validation set at each step, and take the best meta-parameters~$\theta_\mathrm{ACMRR}$ as those corresponding to the lowest recorded validation loss.

\end{document}